\documentclass{article}

\usepackage[nonatbib, final]{neurips_2020}

\usepackage[utf8]{inputenc} 
\usepackage[T1]{fontenc}    
\usepackage{hyperref}       
\usepackage{url}            
\usepackage{booktabs}       
\usepackage{amsfonts}       
\usepackage{nicefrac}       
\usepackage{microtype}      
\usepackage{graphicx}
\usepackage{color}
\usepackage{soul}
\usepackage{amsmath, amsfonts, mathtools, amssymb}
\usepackage{tabulary}
\usepackage[square,numbers]{natbib}

\title{Text Analytics for Resilience-Enabled Extreme Events Reconnaissance}

\author{
  Alicia Y. Tsai \thanks{Correspondence to: \texttt{aliciatsai@berkeley.edu}} \\
  UC Berkeley \\
  \And
  Selim G\"unay \\
  UC Berkeley \\
  \And
  Minjune Hwang \\
  UC Berkeley \\
  \And
  Pengyuan Zhai \\
  UC Berkeley \\
  \And
  Chenglong Li \\
  UC Berkeley \\
  \And
  Laurent El Ghaoui \\
  UC Berkeley \\
  \And
  Khalid M. Mosalam \\
  UC Berkeley \\
}

\begin{document}

\maketitle

\begin{abstract}
Post-hazard reconnaissance for natural disasters (e.g., earthquakes) is important for understanding the performance of the built environment, speeding up the recovery, enhancing resilience and making informed decisions related to current and future hazards. Natural language processing (NLP) is used in this study for the purposes of increasing the accuracy and efficiency of natural hazard reconnaissance through automation. The study particularly focuses on (1) automated data (news and social media) collection hosted by the Pacific Earthquake Engineering Research (PEER) Center server, (2) automatic generation of reconnaissance reports, and (3) use of social media to extract post-hazard information such as the recovery time. Obtained results are encouraging for further development and wider usage of various NLP methods in natural hazard reconnaissance.
\end{abstract}

\section{Introduction}
Due to the exponential growth of Artificial Intelligence (AI) technologies, use of AI methods in structural engineering, similar to many other disciplines, has seen a reasonable increase in recent years. Several studies employed Machine Learning (ML) models to predict the structural responses, fragility parameters or performance limit states of steel and concrete moment frames, steel braced frames and reinforced concrete walls \cite{ML1, ML2}. Other studies used ML models to predict the values of structural model parameters (e.g. shear strength, drift capacity) from experimental data \cite{ML4, ML5}. In the category of image-based structural health monitoring, there has been a reasonable number of ML-based computer vision studies to automatically detect damage from images \cite{ML7, ML8}. Other studies focused on AI-based damage detection, localization and classification using sensors located on structures \cite{ML10, ML11}.

Natural hazards such as earthquakes, tsunamis, and hurricanes, have potential to cause fatalities and injuries as well as damage to buildings and other infrastructure. Post-hazard reconnaissance is therefore important to enhance the understanding the performance of the built environment, speed up the recovery and make informed decisions related to current and future hazards. Although ML-based methods have been used in natural hazards reconnaissance, relatively few studies have explored techniques from natural language processing (NLP). NLP techniques can potentially lead to significant increase of efficiency and accuracy for post-hazard reconnaissance evaluation. In one of the few NLP related work, \citet{ML13} trained a long short-term memory (LSTM) model for classifying building damage using text based natural language damage descriptions according to the green, yellow and red tagging categories of ATC-20. Social media data such as tweets have been used in a few studies to train ML algorithms for direct eyewitness messages in disasters \cite{twitter1} and to identify themes in social media behavior during hazards \cite{twitter2}.

In this paper, two applications of NLP are explored in the context of earthquake reconnaissance. First is the automatic generation of reconnaissance reports, which are an essential part of each field reconnaissance. Automatic report generation aims at decreasing the time to generate a report and increasing the accuracy and abundance of information by facilitating access to many identified resources that can be missed otherwise. Second application is the use of social media and crowd-sourcing to extract information related to earthquake consequences and resilience, such as recovery time, which is difficult to be obtained using other methods. Considering that this is a new untapped application field of AI, the preliminary study conducted herein is expected to lead to the initiation of advances in this area.

Following sections of the paper are organized as follows: The automatic data collection approach used for both applications is described in Section \ref{sec:data}. Methodology and applications of the automatic report generation and recovery time estimation are described in Sections \ref{sec:briefing} and \ref{sec:social-media}. Finally, concluding remarks and future directions are listed in the last section.

\section{Automated Data Collection from News and Social Media Websites}
\label{sec:data}
The automatic data collection is performed by using a Python script that communicates with the U.S. Geological Survey (USGS) Earthquake Hazard Program API (Application Programming Interface) \footnote{\url{https://earthquake.usgs.gov/earthquakes/feed/}}. The program is scheduled to run every day in the Pacific Earthquake Engineering Research (PEER) server and query new earthquakes from the USGS API. Only earthquakes that have magnitude greater than or equal to 5 and USGS PAGER alert level in either yellow, orange or red are recorded. When a new earthquake is detected, the program starts collecting related social media data from Twitter and related news articles from News API \footnote{\url{https://newsapi.org/}}. Tweets are collected over a period of three months using the keyword ``earthquake” and the earthquake location. Tweets are also collected in the local language to capture local effects more precisely. News articles related to the earthquake are collected for duration of a week. The news articles data is then used in the automatic report generation and the social media data is used in the recovery time analysis detailed in the next two sections. 

\section{Automatic Generation of Hazard Briefings}
\label{sec:briefing}
Reconnaissance reports are an essential component of each natural hazard field reconnaissance as they report all findings, observations and conclusions from the event. These reports can be in the form of reports from the detailed field assessments or preliminary reports and briefings based on virtual resources. NLP provides a great venue for automatic generation of hazard briefings as the utilized information is from news websites and websites that provide the characteristics of the hazard (e.g. USGS). Furthermore, briefings provide concise information within well-defined sections. In terms of the hazard type, an event could represent any natural hazard, including earthquake, tsunami, hurricane, etc. However, the focus in this paper is earthquake briefings, therefore the next two subsections explain the methodology developed for automatic generation of earthquake briefings along with the pursued applications. 

\subsection{Methodology}
A typical earthquake briefing consists of standard sections of ``Introduction'', ``Hazard Description'', ``Damage to Buildings'', ``Damage to Other Infrastructure'' and ``Resilience Aspects and Effects on Community''. The ``Introduction'' and ``Hazard Description'' sections include standard contents with only a few items related to the specifics of the event (e.g. date and time, magnitude, location, epicentral coordinates). To complete this specific information in the hazard description, a script is developed that directly communicates with the USGS API and fills out the relevant information automatically. 

The remaining sections are generated using information collected from the new articles. For a given earthquake, the automated data collection script provides us with a set of relevant news articles and their content. In order to generate contents for each remaining sections, ``Damage to Buildings'', ``Damage to Other Infrastructure'' and ``Resilience Aspects and Effects on Community'', we first perform a classification task that classify each sentence in the article into one of the four categories, ''building'', ''infrastructure'', ''resilience'' and ''other''. Sentences that are classified into the first three classes correspond to the sentences that will later be summarized and added to the ''Damage to Buildings'', ''Damage to Other Infrastructure'' and ''Resilience Aspects and Effects on Community'' sections. The fourth class corresponds to any sentence that does not fit into any of the three categories mentioned earlier. The classification task is then followed by the document summarization task. In this task, we employ extractive summarization techniques to condense and summarize all the sentences in each section.

\subsubsection{Sentences Classification}

Five classification methods are considered in the classification task, from simple to complex ones: (1) keyword match, (2) logistic regression (LR), (3) support vector machine (SVM), (4) convolutional neural networks (CNN), and (5) semi-supervised generative adversarial network (GAN). The keyword match method classifies sentences by matching keywords provided by the reconnaissance researchers that are commonly used in the corresponding sections of a briefing.

\paragraph{Training data}
There is currently no available labeled dataset for our application. Therefore, we generate training data using past earthquake briefings and reports that are available. The training dataset contains around 200 sentences collected from the past earthquake briefings and reports and the data is manually labeled by the reconnaissance researchers.

\paragraph{CNN classifier}
We follow the methodology of \citet{kim_CNN}, where the sentences are first embedded by randomly-initialized word vectors, \textit{i.e.}, $w_i\in\mathbb{R}^m, i\in\{1,...,n\}$, where $m$ and $n$ are the embedding size and vocabulary size respectively. Word embedding vectors are updated through back-propagation during training. Three sizes of convolutional filters are used, with heights equal to 3, 4, 5 and width equal to the embedding size. Max pooling is also used, followed by a fully connected layer with softmax activation.

\begin{figure}
    \centering
    \includegraphics[width=.9\textwidth]{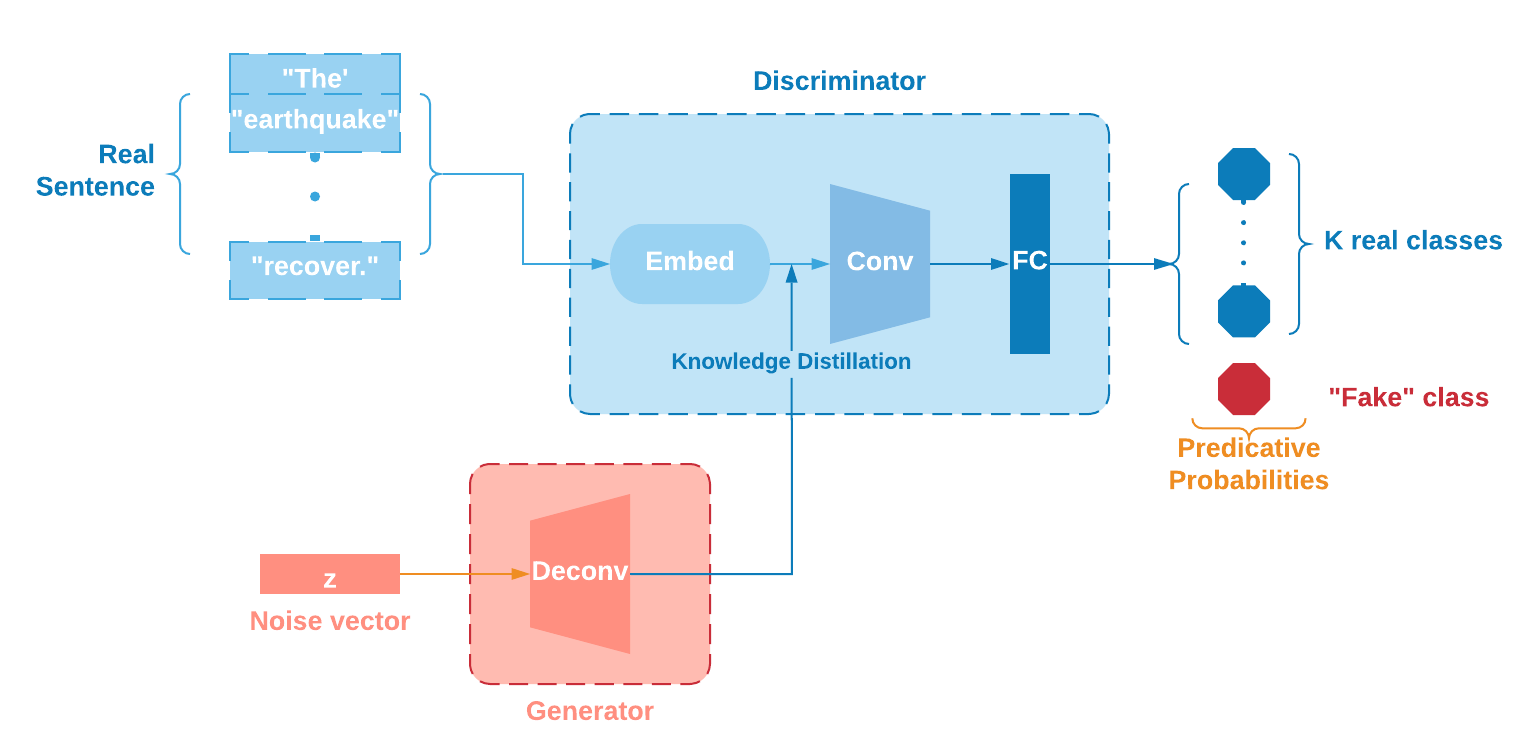}
    \caption{Semi-supervised GAN architecture.}
    \label{fig:TAR_GAN}
\end{figure}

\paragraph{Semi-supervised GAN classifier with Knowledge Distillation}
The Generative Adversarial Network (GAN) is based on the competition between of two parts, the \textit{generator} and the \textit{discriminator}. Our semi-supervised GAN framework is based on \citet{salimans2016improved}, which considers both the supervised and unsupervised losses. The unsupervised loss measures the discriminator's abilities to distinguish ``real'' and ``fake'' data. If a sample is categorized as ``real'', the discriminator also makes a prediction of the sample's class. The supervised loss measures the discriminator's ability to correctly label the real data. Figure \ref{fig:TAR_GAN} illustrates our Semi-supervised GAN framework. Instead of outputting discrete text representations, which prevents the gradient flow through back-propagation, our generator tries to output smooth vector representations in the same manifold of the real text's smooth representations learnt by the discriminator via knowledge distillation \cite{D_knowledge, KD_GAN}. We use the same network architecture for the discriminator as in the CNN classifier, except for the output size of the last fully-connected layer ($K$ outputs for CNN and $K+1$ outputs for GAN). The detailed GAN architecture is summarized in Appendix \ref{appen:gan}.

\subsubsection{Document Summarization}
After the classification task is completed, the second step for the automatic generation of the earthquake briefing is to condense and synthesize sentences in each section. This is accomplished using techniques from document summarization. Automatic document summarization is the process of condensing a set of data computationally in order to create a summary that best represents information of the original content. There are two general approaches to document summarization: \textit{extractive summarization} and \textit{abstractive summarization}. Extractive summarization techniques extract the summary from the original data without modifying the sentences or phrases. In contrast, abstractive summarization may paraphrase the summary. In this work, we consider unsupervised extractive summarization techniques since the briefing generated using NLP is used as a first step to quickly provide useful and relevant information to researchers. Extractive summarization methods generally have more stable performance compared to abstractive methods, since abstractive methods require good performing natural language generation techniques. Furthermore, abstractive methods typically require large amount of labeled training data; however, there is currently no labeled hazard briefings summarization data available. As a result, an unsupervised extractive summarization method called TextRank \cite{mihalcea-tarau-2004-textrank} is used in this work. 

\subsection{Application}
An earlier version of the methodology described above is used to automatically generate summaries for three earthquake briefings for the StEER (Structural Extreme Events Reconnaissance) Network \cite{StEERreport1, StEERreport2, StEERreport3}. In this section, we evaluate the feasibility of the full methodology using Albania earthquake as a case study. We generate an earthquake briefing for the 2019 Mw 6.4 Albania earthquake, which caused significant damage and disruptions to the local community, using the pipeline described above. Around 130 sentences were collected from different news websites. Table \ref{table:classification} summarizes the performance of the classification algorithms on the training data collected from the past briefings and the performance of the algorithms on the news related to Albania earthquake. The results show that keywords matching is quite limited as it is difficult to exhaustively list all of the possible keywords in each section. The other four methods have relatively high performance and therefore in our pipeline we take the majority vote of these four methods as our final classification output. In the case of Albania earthquake, we achieve 69\% classification accuracy with the majority vote. This accuracy is considered sufficient as the automatically generated report is not regarded as a final report, but rather an intermediate document that helps the domain experts create the final document in an accurate and efficient way. Regardless, this accuracy is attributed to the small number of training data used in the algorithms and is expected to increase with larger number of labeled data, which is one of the future agenda of the study. The Albania briefing generated with the full pipeline and the ROUGE score evaluation for the summarization are included in the Appendix \ref{appen:briefing}.

\begin{table}[h]
    \centering
    \caption{Sentence classification accuracy on training data and Albania earthquake test case.}
    \label{table:classification}
    \resizebox{.65\textwidth}{!}{
    \begin{tabulary}{\textwidth}{ ccc }
    \toprule
    \textbf{Classification algorithm} & \textbf{Training data} & \textbf{Albania earthquake} \\
    \midrule
    Keywords & 61\% & 35\% \\
    LR & 100\% & 67\% \\
    SVM & 100\% & 67\% \\
    CNN & 93\% & 75\% \\
    GAN & 88\% & 73\% \\
    \bottomrule
    \end{tabulary}
}
\end{table}

\section{Social Media for Resilience Analysis}
\label{sec:social-media}
In the context of extreme events, recovery time is the time needed after the extreme event to restore the functionality of a structure, an infrastructure system (e.g. water supply, power grid, or transportation network), or a community, to a desired level that can operate or function the same, close to, or better than the condition before the extreme event \cite{resilience1}.

The determination of recovery time using information from social media is based on the assumption that certain keywords related to recovery, (e.g. school, office, transportation, or power outage) appear more frequently on the shared posts, tweets, etc., right after an earthquake occurs and the frequency of these words reduces as time passes. Using this assumption, the time between the occurrence of the earthquake and when these frequencies reduce to pre-earthquake levels is used as a measure of recovery time. The methodology to determine the recovery curve and the recovery time follows the steps below:
(1) determine factors and keywords related to recovery and assign weights to them (schools: 20\%, roads: 20\%, houses: 20\%, offices: 20\%, collapse: 20\%);
(2) determine the variation of the number of posts containing these keywords with time;
(3) determine the recovery time ($t_r$) for each factor from the frequency plots, where $t_r = t_1-t_0$, $t_0$ is the earthquake occurrence time, and $t_1$ is the time when the number of posts with the considered keyword fall below a certain threshold (e.g. 15\% of the maximum frequency) and become steady;
(4) plot the recovery curve considering the weight and recovery time of each factor.

\begin{figure}
    \centering
    \includegraphics[width=.85\textwidth]{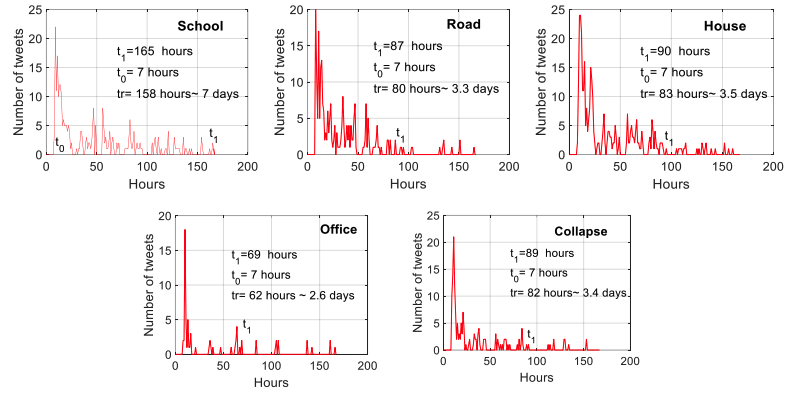}
    \caption{Frequency of social media posts with different keywords over time.}
    \label{fig:weibo}
\end{figure}

A case study is conducted to compute the recovery time by using Weibo posts collected for the Mw 6.6 2013 Ya’an, China
earthquake. The frequency plots for the considered keywords are shown in Figure \ref{fig:weibo}, where $t_1$, $t_0$ and $t_r$ are also specified on each plot. Weighted average of $t_r$ from all factors result in an estimated recovery time of 4 days. The actual recovery time is usually not available or hard to estimate accurately. To better estimate the recovery time, a focused survey was developed and the surveyees were asked if they experienced problems related to several resilience aspects after the earthquake (including access to homes, office shutdown, school closure, power and other utilities outage, transportation issues, interruptions in hospital, retail, telecommunication services, etc.) and how long each lasted. Average responses were considered as the recovery time for each aspect, which were processed in the same way as the social media data to more accurately determine the resulting recovery time and the recovery curve. An example survey is included in the Appendix \ref{append:survey}. Although the focused survey provides much more accurate information, it is generally hard to disseminate these surveys to all the local communities and therefore the number of survey responses can be small compared to the amount of data from social media. Future studies will incorporate events with known recovery times to further explore the approach of combining the information from social media and the high quality information from the focused survey.

\section{Concluding Remarks and Future Directions}
This study explored the usage of text analytics and NLP techniques in natural hazards reconnaissance, particularly for earthquake briefing generation and to extract key information related to earthquake consequences. Results of the conducted study are encouraging for potential advances and more widespread usage of the NLP technologies in natural hazard reconnaissance. Our conclusions are:
(1) reasonable accuracy levels were achieved for the classification of sentences;
(2) briefing sections automatically generated using NLP were similar to the briefings developed by earthquake engineering experts;
(3) it was demonstrated that it is possible to estimate recovery time and curve using information from social media and the focused surveys. 

Future studies and research agenda to this work are considered as follows:
(1) expand and potentially open-source the dataset used for sentence classification to increase accuracy and related research;
(2) include auto-extraction of pictures, photos and descriptions from the news websites in the automatic briefing generation pipeline, and develop methods to potentially use these data in the generation of reports;
(3) perform sentiment analysis for social media data to enhance the recovery time analysis;
(4) open-source the software to encourage further development and wider use of various NLP methods in natural hazard reconnaissance.

\section*{Acknowledgement}
Funding in direct support of this work is provided by Pacific Earthquake Engineering Research Center (PEER) and Berkeley Artificial Intelligence Lab.

\bibliographystyle{abbrvnat}
\bibliography{reference}

\newpage
\appendix

\section{Albania Earthquake Briefing Case Study}
\label{appen:briefing}
\subsection{Example briefing generated by the algorithms}

\paragraph{Damage to Buildings}
A 20-year-old woman, in a coma after she was injured by a falling brick while leaving her apartment in Tirana, died, the health ministry said on Saturday. Rama said on Saturday that preliminary figures showed more than 1,465 buildings in the capital, Tirana, and about 900 in nearby Durres were seriously damaged in Tuesday's 6.4-magnitude predawn earthquake. Rescuers in Albania dug through the rubble of collapsed buildings in search of survivors on Tuesday, after a 6.4-magnitude earthquake struck the Balkan nation, killing at least 23 people and injuring 650. In Durres, hundreds of residents as well as Rama and President Ilir Meta attended the funeral of nine members of a single extended family who were killed when a four-storey villa collapsed. Earthquake damage is being checked by civil engineers from the European Union, United States and local experts to assess whether buildings are structurally sound, unsafe and required demolition or just needed replastering. He said more than 1,465 buildings in Tirana and about 900 in nearby Durres had been seriously damaged. Durres castle walls damaged by the earthquake In Albania, a large proportion of the earthquake damage has been blamed on corruption, violations of the building code and substandard construction following the demise of communism during the early 1990s. About 2,500 people from damaged homes have been sheltered in hotels. The earthquake struck at 3:54 a.m. near the Adriatic coast, about 19 miles west of Tirana, home to nearly 900,000 people. Four buildings, including a five-storey apartment block, collapsed in Kodër-Thumanë and the town was hardest hit from the earthquake. Of those, more than 3,000 people were injured, 14,000 became homeless and throughout Albania 14,000 buildings were damaged of which 2,500 are rendered uninhabitable. In Elbasan, a town about 35 miles from Durres, Olsi Shehi, a 39-year-old cook, said a four-story house had fallen, trapping people inside. Everything was moving in an unbelievable rhythm, I could hear the walls cracking, dishes and glass breaking. Videos and pictures shared on social media showed chaotic scenes of residents rummaging through the rubble, or trying to extricate people trapped under collapsed buildings.

\paragraph{Damage to Other Infrastructure}
In early February 2020, the Albanian government publicised figures that earthquake damage to private and public properties cost a 844 million.

\paragraph{Resilience Aspects and Effects on Community}
In the immediate aftermath, 2,500 people became displaced by the earthquake and are temporarily being accommodated either in the Niko Dovana Stadium of Durres in tents or in hotels. A state of emergency lasting 30 days was declared by the Albanian government for Durres, Thumanë and Tirana and later extended to Lezhë and LaAS. Subsequently, rescue crews with specialised equipment, sniffer dogs and emergency supplies came to Albania from neighbouring countries and other European nations to help in the search efforts and provide for those left homeless. Prime Minister Rama said that the state budget was being reconfigured to manage the situation following the earthquake. Blue and white coloured emergency tents for displaced people near stadium in Durres The EU office in Albania estimated that some 1.9 million people out of a total population of 2.8 million have been affected by the earthquake. The search-and-rescue operation for earthquake survivors in Albania has ended, with the death toll standing at 51 and no more bodies believed to be in the ruins, Prime Minister Edi Rama said. On 30 November Prime Minister Rama announced the end of the search and rescue operation, as no more bodies were expected to be under the rubble. Some students from Tirana went to assist relief efforts in Durres and delivered hundreds of meals to earthquake affected people. Hundreds of Albanians in Albania and Kosovo opened their homes to people displaced by the earthquake. In accordance with the Albanian constitution regarding an emergency situation, the Albanian parliament granted Prime Minister Edi Rama state of emergency powers to deal with earthquake aftermath.

\begin{table}[h]
    \centering
    \caption{ROUGE F1-score for summarization results of each section}
    \label{table:rouge}
    \begin{tabular}{ cccc }
    \toprule
    \textbf{Section} & \textbf{ROUGE-1} & \textbf{ROUGE-2} & \textbf{ROUGE-L} \\
    \midrule
    Damage to Buildings & 33.3 & 7.4 & 21.4 \\ 
    Damage to Other Infrastructure & 13.3 & 0.0 & 12.5\\ 
    Resilience Aspects and Effects on Community & 39.9 & 14.7 & 31.4 \\
    \bottomrule
    \end{tabular}
\end{table}

\subsection{Albania earthquake briefing generated by domain experts}
\paragraph{Damage to Buildings} 
Albanian Prime Minister Edi Rama indicated that more than 1,465 buildings in Tirana and about 900 in the nearby city Durres had been seriously damaged. Many reinforced concrete (RC) and masonry buildings experienced collapse and severe damage. Two hotels and two apartment blocks collapsed in Durres. Four buildings, including a five-story apartment block, collapsed in Thumane. At the time this briefing was authored, many people were still trapped in the remains of the ruined buildings. An illustration of insufficient detailing from a collapsed RC building is provided by figure: the exposed vertical element shows a lack of transverse reinforcement and failure in a diagonal plane associated with shear damage; the horizontal element in this figure is observed to have adequately anchored transverse reinforcement with only 90-degree hooks in place of the 135-degree seismic hooks necessary for confinement. These and other non-ductile features such as the presence of strong-beam/weak-column proportions, lack of confinement at member ends and connections and weak/soft stories potentially contributed to the observed collapses. One of the commonly observed damage types of RC buildings in this earthquake is the In-Plane (IP)/Out-of-Plane (OOP) failures of infill walls. Fortunately, the damage in these photos was limited to the infill failures and did not result in the formation of weak and soft stories and consequent story collapses. However, infill wall failures may have contributed to other building collapses, similar to those observed during previous earthquakes in Europe. Infill wall failures also present a high risk of injury or death due to falling masonry rubble. Related to the damage in masonry buildings, photos show the presence of multicell clay blocks. These blocks are not only very brittle but they afford no options for reinforcing or grouting the cells to increase the wall strength and ductility. The use of such brittle material should be outlawed in all earthquake-prone areas, including Albania and other countries around the region.

\paragraph{Damage to Other Infrastructure}
Because the earthquake caused significant building damage, collapses and consequent fatalities, almost all of the preliminary information is on buildings. At the time this briefing was authored, there was not much information available related to the damage of other infrastructure, though the photo shows significant road damage in the capital city Tirana.

\paragraph{Resilience Aspects and Effects on Community}
USGS PAGER tool estimated the fatalities to be between 1 and 10 with a probability of 12\%, between 10 and 100 with a probability of 37\%, between 100 and 1,000 with a probability of 37\% and between 1,000 and 10,000 with a probability of 12\%. At the time this briefing was authored, the number of deaths as a consequence of the earthquake was reported as 51 and there were approximately 2,000 injuries. Damages were expected to be between \$1 million and \$10 million, between \$10 million and \$100 million, and between \$100 million and \$1,000 million with probabilities of 8\%, 25\% and 36\%, respectively. Furthermore, there were probabilities of 22\% and 6\% of the economic loss to be between \$1,000 million and \$10,000 million and between \$10,000 million and \$100,000 million, respectively. Given the severity of the situation, Albanian Prime Minister Edi Rama declared a state of emergency in Tirana and Durres during December. Recovery efforts are currently continuing in the rubble of collapsed buildings, where residents and emergency crews in cities across the country rescued 45 people from some of the collapsed buildings. Considering the state of emergency and current situation, the recovery and reconstruction process after this earthquake is likely to be lengthy. The earthquake left around 4,000 people homeless. Similar to many previous earthquakes, even the residents of houses and buildings that were still standing, which performed well, remained outside after the earthquake. One of the residents in the capital Tirana indicated he did not know where he would live and described his apartment as “uninhabitable.” An estimated 2,500 people have been displaced by the earthquake and are temporarily being sheltered either in the Niko Dovana Stadium of Durres in tents or in hotels.

\newpage
\section{Focused Survey Questions for Recovery Time}
\label{append:survey}
\begin{enumerate}
    \item Where are you located? City/State/Country
    \item Was your house/building yellow or red tagged? 
    \item If yes, what was the duration that you could not enter your home?
    \item Did you have power outage at home? 
    \item If yes, how many hours, days or weeks did it last?
    \item Did you have interruption of any other utility services, gas, water, etc?
    \item If yes, how many hours, days or weeks did they last?
    \item Was your office shut down?
    \item If yes, how many hours, days or weeks did it last?
    \item Was your or your children’s  school closed?
    \item If yes, how many hours, days or weeks did it last?
    \item Did you have any increase in your commute time after the earthquake?
    \item If yes, how much was the increase (in \%)?
    \item If yes, how many hours, days or weeks did the increased commute time last?
    \item Was there any slow down in the hospital services?
    \item If yes, how many hours, days or weeks did it last?
    \item Was there any interruption in the retail store services, restaurants, supermarkets, department stores, banks, etc? 
    \item If yes, how many hours, days or weeks did it last?
    \item Was there any interruption in the telecommunication (e.g. cellphone) services? 
    \item If yes, how many hours, days or weeks did it last?
    \item Was there any interruption in the internet provider services? 
    \item If yes, how many hours, days or weeks did it last?
    \item Is your daily life impacted by the hazard? 
    \item If yes, how many hours, days or weeks did the impact last?
    \item If yes, please describe how your daily life is impacted by the hazard.
    \item Did you experience any other consequences not covered by the questions above?
\end{enumerate}

\section{Semi-supervised GAN}
\label{appen:gan}
We summarize the GAN training procedure as such: 

\begin{enumerate}
    \item For a $K$ real class dataset, given a real sentence $s=\{w_1, w_2,...,w_n\}$ with real label $k\in{1,...,K}$, for each word $w_i$, the discriminator embedding layer outputs a vector $v_i = \big[v^{(i)}_1, \cdots, v^{(i)}_m \big]^T$, where $m$ is the embedding size. An embedding matrix $M = \big[v_i, \cdots, v_n \big]^T, M\ in \mathbb{R}^{n\times m}$ is fed into subsequent convolutional and max-pooling layers, and finally reaches the fully-connected layer, which outputs a logit vector for this sentence sample $l = \big[l_i, \cdots, l_{K+1} \big]$, where $l_{K+1}$ signifies the logit for the ``fake'' class. The predicative probabilities are calculated by the softmax function, \textit{i.e.}, $p(\hat{y} = i \,| \, s) = \dfrac{\exp(l_i)}{\sum_{j=1}^{K+1}\exp(l_j)}, i \in \big\{1, \cdots, K+1 \big\}$. 
    
    \item Randomly sample a latent noise vector $z$ from the standard Gaussian distribution with an arbitrary length (e.g., 100) and feed it into the deconvolutional layers in the Generator, which outputs a ``fake'' sentence embedding matrix $M_{\text{fake}} \in \mathbb{R}^{n\times m}$. Feed $M_{\text{fake}}$ directly into the convolutional layers of the discriminator, bypassing the embedding layer. The discriminator outputs the predictive probabilities $p(\hat{y} = i \,| \, M_{\text{fake}}), i \in \big\{1, \cdots, K+1 \big\}$. 
    
    \item Update the discriminator and generator parameters through back-propagation using $L^{(D)}$ and $L^{(G)}$ (Eq.\ref{eq:D_loss} and Eq. \ref{eq:G_loss})
\end{enumerate}

In the case where real data are labeled, the Discriminator also makes a prediction on individual classes on top of predicting the data as ''real'' or ''fake''. The loss function defined as followed:

\begin{align}
    L^{(D)} = & L^{(D)}_{\text{unsupervised}} + L^{(D)}_{\text{supervised}}
    \label{eq:D_loss} \\
    L^{(D)}_{\text{unsupervised}} = & - E_{x \sim P_{\text{data}}} \log(1 - p_{\text{model}}(y=K+1 \,| \, x)) \\
    & - E_{x \sim P_{\text{generator}}} \log(p_{\text{model}}(y = K+1 \,| \, x))
    \label{eq:unsupervised_loss} \\
    L^{(D)}_{\text{supervised}} = & - E_{x,y \sim P_{\text{data}}(x,y)} \log(p_{\text{model}}(y \,| \, x, y < K+1))
    \label{eq:supervised_loss}
\end{align}

where $K$ is the total number of real classes (3 in our case for ``building,'' ``infrastructure,'' and ``resilience'') and $K+1$ denotes the ``fake'' class in which, all data are synthesized by the generator. $L^{(D)}_{\text{unsupervised}}$ measures the discriminator's abilities to recognize ``real'' and ``fake'' data (corresponding to the first and second term respectively). $L^{(D)}_{\text{supervised}}$ measures the discriminator's ability to correctly label data in their respective ``real'' classes (hence the $y < K+1$ condition).

The generator's goal is to ``undermine'' the discriminator's performance by creating realistic-looking data. This goal can be reflected in two parts of the generator loss function: the game loss, and the feature matching loss.
\begin{align}
    L^{(G)} = & L^{(G)}_{\text{game}} + L^{(G)}_{\text{feature-matching}}
    \label{eq:G_loss} \\
    L^{(G)}_{\text{game}} = & -E_{z \sim p_z} log[1 - p_{\text{model}}(y = K+1 \,| \, G(z))]
    \label{eq:SSGAN_heristic_loss} \\
    L^{(G)}_{\text{feature-matching}} = & \Big\Vert E_{x \sim p_{\text{data}}(x)} f(x)-E_{z\sim p_{z}(z)} f(G(z)) \Big\Vert_{2}^{2}
    \label{eq:G_loss_fm}
\end{align}

The $L^{(G)}_{\text{game}}$ term is a direct reflection of ``how bad'' the discriminator is when facing a realistic, generated sample (by mislabeling it as ``real''). The $L^{(G)}_{\text{feature-matching}}$ term measures the similarities between the intermediate layer activations ($f(x)$) between real and fake data (we choose $f(x)$ to be the ReLU-activated feature map output by the last convolutaional layer in the Discriminator).

\begin{table}
\centering
    \caption{Configurations of the discriminator and generator of the Semi-supervised GAN}
    \label{tab:GAN-config}
    \begin{tabulary}{\textwidth}{cccc}
    \toprule
    \multicolumn{4}{c}{\textbf{Discriminator}} \\
    \midrule\midrule
    \textbf{Layer} & \textbf{Precedent Layer} & \textbf{Activation} & \textbf{Output Shape} \\
    \midrule
    Real Input     &         -         &        -       &   ($N$, 64)  \\
    Real Embed      &   Real Input  &    -  &   ($N$, 64, 80) \\
    Fake Embed     &         -         &        -       &   ($N$, 64, 80)  \\
    Conv-1 3 filters size=3$\times$80$\times$1& Real or Fake Embed     &    Leaky ReLU  &   ($N$, 62, 1, 3) \\
    MaxPool-1 kernel size=62$\times$1$\times$1 & Conv-1                &        -       &   ($N$, 1, 1, 3) \\
    Conv-2 3 filters size=4$\times$80$\times$1& Real or Fake Embed     &    Leaky ReLU  &   ($N$, 61, 1, 3) \\
    MaxPool-2 kernel size=61$\times$1$\times$1 & Conv-2                 &        -       &   ($N$, 1, 1, 3) \\
    Conv-3 3 filters size=5$\times$80$\times$1& Real or Fake Embed     &    Leaky ReLU  &   ($N$, 60, 1, 3) \\
    MaxPool-3 kernel size=60$\times$1$\times$1 & Conv-3             &        -       &   ($N$, 1, 1, 3) \\
    Concat  &  MaxPool-1,2,3     &- & ($N$, 1, 1, 9) \\
    Dropout (rate = 0.25)   &        Concat         &        -       &   ($N$, 1, 1, 9) \\
    Flatten   &         Dropout         &        -       &   ($N$, 9) \\
    Fc-layer  &         -         &     Softmax    &   ($N$, $K+1$) \\
    \bottomrule 
    \end{tabulary}
    
    \vspace{1em}
    
    \begin{tabulary}{\textwidth}{ccccc}
    \toprule
    \multicolumn{5}{c}{\textbf{Generator}} \\
    \midrule\midrule
    \textbf{Layer} & \textbf{Filter size (\#)} & \textbf{Activation} & \textbf{Output Shape} & \textbf{Notes} \\
    \midrule
    Input       &         -          &        -       &   ($N$, 100)              &    From Normal distribution   \\
    Fc-layer    &         -          &       ReLU     &   ($N$, 40,960)          &    40,960 = 16$\times$20$\times$128    \\
    Reshape     &         -          &        -       &   ($N$, 16, 20, 128)      &     - \\
    Deconv      &   3$\times$3 (128)  &       ReLU     &   ($N$, 32, 40, 64)       &    Stride = 2 \\
    BatchNorm   &         -          &        -       &   ($N$, 32, 40, 64)       &    Momentum = 0.8 \\
    Deconv      &   3$\times$3 (64)   &       ReLU     &   ($N$, 64, 80, 3)      &    Stride = 2 \\
    BatchNorm   &         -          &        -       &   ($N$, 64, 80, 3)      &    Momentum = 0.8 \\
    Deconv      &   3$\times$3 (3)    &       -     &   ($N$, 64, 80, 3)      &    Stride = 1 \\
    \bottomrule
    \end{tabulary}
\end{table}

\end{document}